\documentclass[runningheads]{llncs}
\usepackage[T1]{fontenc}

\usepackage{bm}
\usepackage{graphicx}
\usepackage{url} 
\usepackage{subfig}
\usepackage{hyperref}

\begin{document}
\title{FusionNet: a frame interpolation network for 4D heart models %
\thanks{ {\bf This is the author's version of the paper. The final authenticated publication is available online at} \protect\url{https://doi.org/10.1007/978-3-031-47425-5_4}}}
%
%
\author{Chujie Chang\inst{1} \and
Shoko Miyauchi\inst{1} \and
Ken'ichi Morooka\inst{2} \and
Ryo Kurazume\inst{1} \and
Oscar Martinez Mozos\inst{3}}
\authorrunning{C. Chang et al.}
%
\institute{Kyushu University, Fukuoka 819-0395, Japan\\
\email{miyauchi@ait.kyushu-u.ac.jp} \and
Okayama University, Okayama 700-8530, Japan \and
\"{O}rebro University, \"{O}rebro 701-82, Sweden
}
\maketitle              
\begin{abstract}
Cardiac magnetic resonance (CMR) imaging is widely used to visualise cardiac motion and diagnose heart disease. 
However, standard CMR imaging requires patients to lie still in a confined space inside a loud machine for 40-60 min, which increases patient discomfort. 
In addition, shorter scan times decrease either or both the temporal and spatial resolutions of cardiac motion, and thus, the diagnostic accuracy of the procedure. 
Of these, we focus on reduced temporal resolution and propose a neural network called FusionNet to obtain four-dimensional (4D) cardiac motion with high temporal resolution from CMR images captured in a short period of time. The model estimates intermediate 3D heart shapes based on adjacent shapes. 
The results of an experimental evaluation of the proposed FusionNet model showed that it achieved a performance of over 0.897 in terms of the Dice coefficient, confirming that it can recover shapes more precisely than existing methods.
This code is available at: \url{https://github.com/smiyauchi199/FusionNet.git}

\end{abstract}
\section{Introduction}
Cardiac magnetic resonance (CMR) imaging provides non-invasive 4D (i.e. 3D space + time) visualisation (cine images in Figure \ref{fig:FusionNet}) of a beating heart over a cardiac cycle. 
The cardiac cycle consists of two periods: diastole, during which the heart muscle relaxes, and systole, during which the heart contracts.
By observing the cardiac motion during this cycle, physicians can diagnose diseases and cardiac defects \cite{1}. 
However, CMR scans are time-consuming and require patients to lie still in a confined space inside a loud machine for 40-60 minutes, which increases patient discomfort. 
Moreover, reducing the acquisition time lowers either or both the temporal and spatial resolutions of the resultant cardiac images, which reduces their usefulness to physicians in accurately diagnosing heart disease.

To reduce CMR scanning times while maintaining spatial and temporal resolutions, various studies \cite{2}\cite{3} have been conducted on the application of neural network models. 
By contrast, in studies on the temporal resolution of CMR images, Lyu et al. \cite{4} proposed a recurrent neural network (RNN) with a bi-directional convolutional long short-term memory (ConvLSTM) architecture to reduce motion artifacts and perform frame interpolation tasks.
However, their method assumed that only one frame of the CMR image was missing from a sequence and did not consider the overall improvement in temporal resolution.
In addition, the inputs for their method were 3D (i.e. 2D space + time) CMR images. 
Kalluri et al. \cite{5} proposed a frame-interpolation network designed to utilise spatiotemporal convolutions to predict the intermediate frames of a video. 
However, this method is intended for 3D (i.e. 2D space + time) media, and was not designed for application to 4D media, such as 4D CMR images.
The frame interpolation results for 4D CRM images can be obtained by stacking the output of each slice obtained using these methods. 
However, in this case, smooth interpolation between slices is difficult because the interpolation of each slice is performed independently.

Therefore, to recover 4D high-frame-rate heart motion from 4D low-frame-rate heart motion generated from CMR images simultaneously, we propose a new frame interpolation network called FusionNet.
Furthermore, the results of an experimental evaluation of the performance of the proposed FusionNet confirmed its effectiveness compared with existing methods.

\section{Dataset}
\label{sec:GHM}
In this study, we represent one cardiac cycle as a set of 3D voxel models covering the one cardiac cycle in time, called a 4D heart model. 
Each 3D voxel model represents a 3D heart shape at one specific frame.
Among CMR images, 4D cine images from the UK Biobank Resource under Application No. 20209 were used to generate 3D heart models, each corresponding to a single cardiac cycle. 
The details of the cine images can be found in the UK Biobank \cite{ukbiobank_cine_images}.
The original cine image dataset contains 50 frames to represent each cardiac cycle.
To reduce the complexity of the problem, we only use 10 frames that were subsampled every 5 frames from the original 50 frames.
We sampled equally because we want to keep one full cardiac cycle. 

In the generation of the 4D heart models, to segment the myocardial region of the left ventricle in each frame, a joint learning model \cite{6} is first applied to a sequence (a set of ten frames) of cine images.
Subsequently, a voxel model consisting of 80 $\times$ 80 $\times$ 80 voxels representing the 3D shape of the segmented myocardial region is generated for the sequence in a similar way to \cite{7}.
In the voxel model, myocardial regions are represented by 1 and other regions by 0.
This set of ten voxel models is called a high-frame-rate (HFR) heart model $\bm X_h$ ($=80 \times 80 \times 80 \times 10$ voxels) which is used as the ground truth. 
In all heart models, the first frame corresponds to the end-diastole of the cardiac cycle. 
In addition, the heart models are spatially aligned based on the 3D shape of the end-diastole.
The HFR set is subsampled at different intervals (lower frequencies) to form a low-frame-rate (LFR) heart model.
In our experiments, the subsampling frame interval was set to one, resulting in a subset of five voxel models corresponding to the odd-numbered frames used as the LFR heart model $\bm X_l$ ($=80 \times 80 \times 80 \times 5$ voxels).

\section{FusionNet architecture}
\label{sec:fusionnet}
The input to FusionNet is the LFR heart model $\bm X_l$, and the output is its corresponding HFR heart model $\bm X_h$.

Various diagnostic support systems \cite{7} treat as input only the segmentation results and the volume changes in the heart region calculated from them, not the original MR images.
Therefore, FusionNet outputs directly the set of voxel models generated from the segmentation results instead of the original MR images.

\begin{figure}[tb]
\begin{center}
\includegraphics[width=\textwidth]{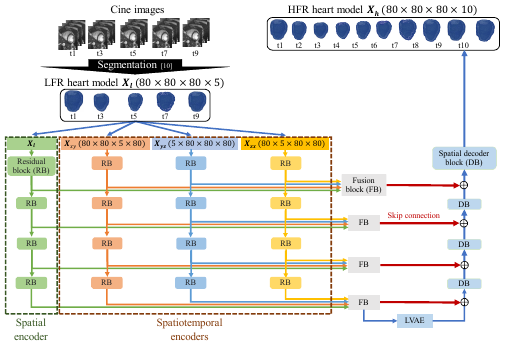}
\caption{Overview of our FusionNet.}
\label{fig:FusionNet}
\end{center}
\end{figure}

As shown in Figure \ref{fig:FusionNet}, the architecture of the proposed FusionNet is constructed by adding skip connections, residual blocks \cite{8}, and spatiotemporal encoders \cite{5} to a baseline. 
The baseline is constructed using a generative model \cite{7} which uses the 4D heart models as the input and generates the 4D heart models for visualisation of the shape features that contributed to a classification.
Similar to the generative model, our study also aims to build a network with the heart models as the input and output. 
Therefore, we employed the generative model as the baseline of our network.

The generative model \cite{7} is an explainable classification network for hypertrophic cardiomyopathy. 
It is composed of a spatial convolutional autoencoder, a three-layer ladder variational autoencoder (LVAE), and a multilayer perceptron (MLP).
The input of the network is the 4D heart model containing two-frame voxel models, and the output is also the heart model containing two-frame voxel models restored by LVAE and the classification result obtained by MLP.
To achieve frame interpolation from the LFR to the HFR heart model, we modified the baseline network by removing the MLP and changing the number of input and output frames in a 3D convolutional autoencoder.
Based on the results of the preliminary experiments, we set the latent variables for each layer of the LVAE to 64, 48, and 32 dimensions.

FusionNet comprises three additional elements compared to the baseline.
The first is the addition of skip connections to the spatial encoder and decoder layers of the baseline to prevent the loss of pixel details in the generated image.

The second is the addition of residual blocks \cite{8} to the spatial encoder to solve the degradation problem, which commonly affects deep networks.
The residual block consists of two paths: main and skip connection paths.
In our network, the main path comprises two convolution blocks, each of which has a 3 $\times$ 3 $\times$ 3 kernel, whereas the skip connection path has one convolution block of the same kernel size.

The third is the addition of spatiotemporal encoders \cite{5} to extract the features of changes in the heart shape over time.
In the baseline spatial encoder, 3D convolution is performed for the 3D space of the heart model.
However, the spatial encoder does not consider convolution for temporal shape changes.
By contrast, the spatiotemporal encoder performs 3D convolution in 2D space and time. Therefore, it extracts features while considering the temporal 2D shape changes.
Given that the heart model contains three-dimensional spatial information, three types of spatiotemporal encoders are trained by transposing the axes of $\bm X_l$: $\bm X_{xy} = 80 \times 80 \times 5 \times 80, \bm X_{yz} = 5 \times 80 \times 80 \times 80, \bm X_{zx} = 80 \times 5 \times 80 \times 80$.
This changes the combination of 2D spaces to be convolved for each spatiotemporal encoder.
To achieve 4D convolution, we combined a spatial encoder and three types of spatiotemporal encoders.
The spatial encoder and three spatiotemporal encoders have the same structure.
Each encoder consists of four residual blocks, and the output feature from each residual block is fed into the fusion block.

In our network, a fusion block is constructed based on a gated information fusion (GIF) block \cite{9}. 
The GIF block can adaptively fuse different feature maps obtained from multimodal inputs by learning different weights for each feature map.
As shown in Figure \ref{fig:FusionNet}, in our fusion block, the four feature maps obtained from the four encoders are concatenated as a feature map.
Next, a convolution block with a 3 $\times$ 3 $\times$ 3 kernel is applied to the concatenated feature map, and a sigmoid function is applied to produce a weight map. 
Finally, the weight map is multiplied by the concatenated feature map, and a convolution block with a 3 $\times$ 3 $\times$ 1 kernel is applied to the multiplication result.

In the FusionNet inference, as shown in Figure \ref{fig:FusionNet}, an LFR heart model is input to each encoder which consists of four residual blocks.
The output of the residual block at a given depth of each encoder is input to the fusion block at that depth to obtain a fused feature of the four encoders.
Using skip connections, the fused feature of the deepest fusion block is input to the encoder of the LVAE, whereas the other fused features are input to the corresponding spatial decoding block of the spatial decoder.
The output of the LVAE decoder is then input into the spatial decoder. 
At each spatial decoding block of the spatial decoder, the input is concatenated with the corresponding fused feature.
Finally, the spatial decoder outputs the HFR heart model.

The loss function of FusionNet comprises the following four terms:
1) A Dice loss term $DL$ is used to evaluate the degree of similarity between the generated heart model and the ground truth. 
2) Three Kullback-Leibler divergence terms $KL_{i}$ ($i = 1, 2, 3$) are used to penalise the deviations between the prior and posterior distributions at each level of the three-layer LVAE. 
As the prior distribution for the highest level, we set the standard Gaussian $\mathcal{N}\left(0, 1\right)$.
Therefore, the loss function $L_{system}$ can be expressed as follows:
$
L_{system} = DL + \alpha \sum_{i=1}^{3} \beta_{i} {KL}_{i},
$
where $\alpha$ and $\beta_{i}$ are the weighting factors for each term.
In this study, $\alpha$, $\beta_1$, $\beta_2$, and $\beta_3$ were set to 1.0, 0.001, 0.001, and 0.01, respectively.

FusionNet is trained to minimise $L_{system}$.
The batch size and number of epochs were set to 10 and 500 in our experiments.
Early stopping was introduced, and the Adam optimiser was used.
The learning rate was set to $1\times e^{-3}$ and multiplied by 0.5 for each set of 30 epochs.

\section{Experiment and discussion}
To evaluate the effectiveness of FusionNet, its accuracy was compared with that of existing methods.
We then conducted an ablation study on different components of the FusionNet architecture to evaluate their specific contributions.
In the experiments, we simulated frame interpolation over a cardiac cycle (from 1 to 10 frames) to generate an HFR heart model composed of ten voxel models from an LFR heart model composed of five voxel models in odd-numbered frames (1, 3, 5, 7, and 9).
The FusionNet was trained from scratch.
In the experiments, statistical significance was tested using the $t$-test.

Each HFR model was divided into two subsets using subsampling. 
The first subset (input) comprised odd-numbered frames (1, 3, 5, 7, and 9) and corresponded to the input LFR set. 
The second set (estimated) comprised of even-numbered frames (2, 4, 6, 8, and 10). 
The estimated set was used as the ground truth. 
In our experiments, we used FusionNet to estimate frames 2, 4, 6, 8, and 10 using LFR frames 1, 3, 5, 7, and 9 as inputs. 

Because it is generally unknown whether a subject is healthy or diseased at the time of scanning, experiments were conducted on datasets that included healthy and diseased subjects without distinction.
The original dataset comprised CMR images obtained from 210 subjects (100 patients with ischemic heart disease and 110 healthy subjects). 
First, 210 heart models were generated from CMR images, as described in Section \ref{sec:GHM}.
Then, to perform 7-fold cross-validation, the heart models were repeatedly divided into 150 samples as a training dataset, 30 samples as a validation dataset, and 30 samples as a testing dataset while varying the subjects included in the validation and testing datasets.
In addition, we performed data augmentation by shifting and rotating the heart models in the training dataset, increasing the number of heart models in the training dataset from 150 to 1200.

To measure the accuracy of our approach, we calculated the Dice similarity coefficient \cite{12} between each generated heart 3D voxel model and the corresponding ground-truth 3D voxel model. 

\begin{table}[tb]
\begin{center}
\caption{Comparison with previous methods using the average Dice coefficient from 7-fold cross validation.}
\label{table:comparison}
\scalebox{0.9}[0.9]{
\begin{tabular}{c|ccccc|c}
\hline
		& Frame 2	&	Frame 4	&	Frame 6	&	Frame 8	&	Frame 10 &	Average 	\\ \hline
FusionNet	&	\bf{.897 $\pm$ .005}	&	\bf{.879 $\pm$ .006}	&	\bf{.877 $\pm$ .007}	&	\bf{.928 $\pm$ .003}	&	\bf{.905 $\pm$ .006}	&	\bf{.897 $\pm$  .019}	\\
													
ConvLSTM \cite{4}	&	.884 $\pm$ .003	&	.867 $\pm$ .005	&	.854 $\pm$ .006	&	.912 $\pm$ .003	&	.892 $\pm$ .007	&	.881 $\pm$ .020	\\
													
U-Net \cite{11}	&	.892 $\pm$ .004	&	.875 $\pm$ .004	&	.871 $\pm$ .007	&	.922 $\pm$ .003	&	.899 $\pm$ .008	&	.892 $\pm$ .018	\\
													
Bilinear \cite{13}	&	.821 $\pm$ .008	&	.831 $\pm$ .008	&	.813 $\pm$ .008	&	.915 $\pm$ .004	&	.890 $\pm$ .005	&	.854 $\pm$ .041	\\ \hline

\end{tabular}
}
\end{center}
\end{table}

\subsection{Comparison with existing methods}
There are various methods based on neural networks for interpolating the frames of 3D (2D space and time) images.
Among them, in this experiment, we compared FusionNet with three existing methods: a ConvLSTM-based method \cite{4}, a U-Net-based method, and a bilinear interpolation. 

The ConvLSTM-based method is a state-of-the-art frame interpolation method for 3D (2D space and time) cine images.
In the method, the voxel model of the estimated frame is obtained using voxel models of multiple adjacent frames centred on the missing frame. In our experiments, the number of input adjacent voxel models was set to four. For example, if we aim to estimate frame number 4, we put two voxel models for frames 1 and 3 into the forward ConvLSTM branch and used two voxel models for frames 5 and 7 in the backward ConvLSTM.
Using a separately trained network for each target frame number, we obtained an HFR heart model from the corresponding LFR heart model.

U-Net \cite{11} is frequently used in medical image processing.
In this study, the U-Net-based method was constructed by replacing the autoencoder of the U-Net with a 3D convolutional autoencoder for 3D space.
The input and output of the U-Net-based method were the same as those of FusionNet; that is, given an LFR heart model as the input, the HFR heart model was the output in a single inference.

The bilinear interpolation is a simple traditional method that does not use neural networks.
In the interpolation, a voxel model of the estimated frame is calculated using the voxel models of two adjacent frames.
When a simple linear interpolation was applied, the interpolation accuracy was low.
To improve the accuracy, for each target voxel, we calculated the target voxel value by considering neighboring voxel values based on the idea of a traditional bilinear interpolation \cite{13}.

\begin{figure}[tb]
\begin{center}
\includegraphics[width=0.9\textwidth]{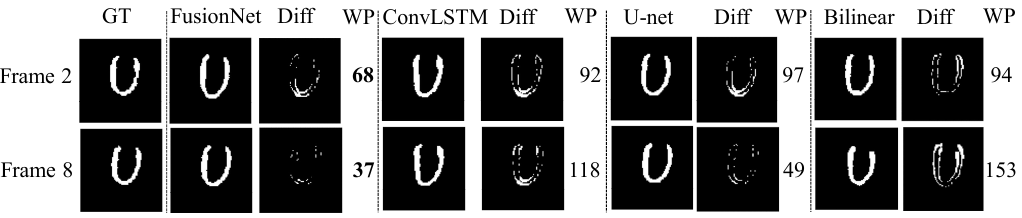}
\caption{Cross-section of the generated voxel model corresponding to the estimated frame (e.g. frames 2 and 8) and the difference image between the model and the ground truth. The last column, WP, indicates the number of white pixels in the difference image. The results of our FusionNet contain the lower values.}
\label{fig:figure1_comparison}
\end{center}
\end{figure}

Table \ref{table:comparison} presents the results of the 7-fold cross-validation. 
Each value in the table shows the average Dice coefficient of FusionNet for the estimated frames of the testing dataset compared with that of the existing methods. 
FusionNet performed better than the three existing methods for all the frames.
For the average Dice coefficient, FusionNet performed better, with a statistical significance level of 0.05. 
In addition, the first column in Figure \ref{fig:figure1_comparison} shows the cross-section of the voxel model corresponding to the ground truth. The following columns show the pixel differences from the models estimated using the four approaches. 
The final column lists the number of different pixels. Our FusionNet achieved a lower pixel difference.
Overall, these results show that the proposed FusionNet can generate HFR heart models with higher accuracy than the existing methods.

\begin{figure}[tb]
\begin{center}
\subfloat[][]{\includegraphics[width=0.38\textwidth]{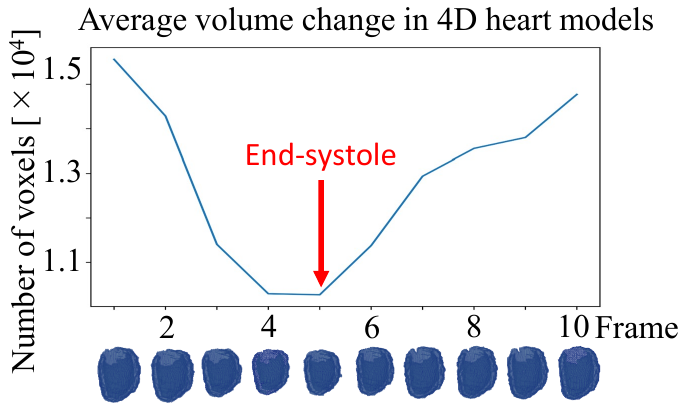}}
\subfloat[][]{\includegraphics[width=0.62\textwidth]{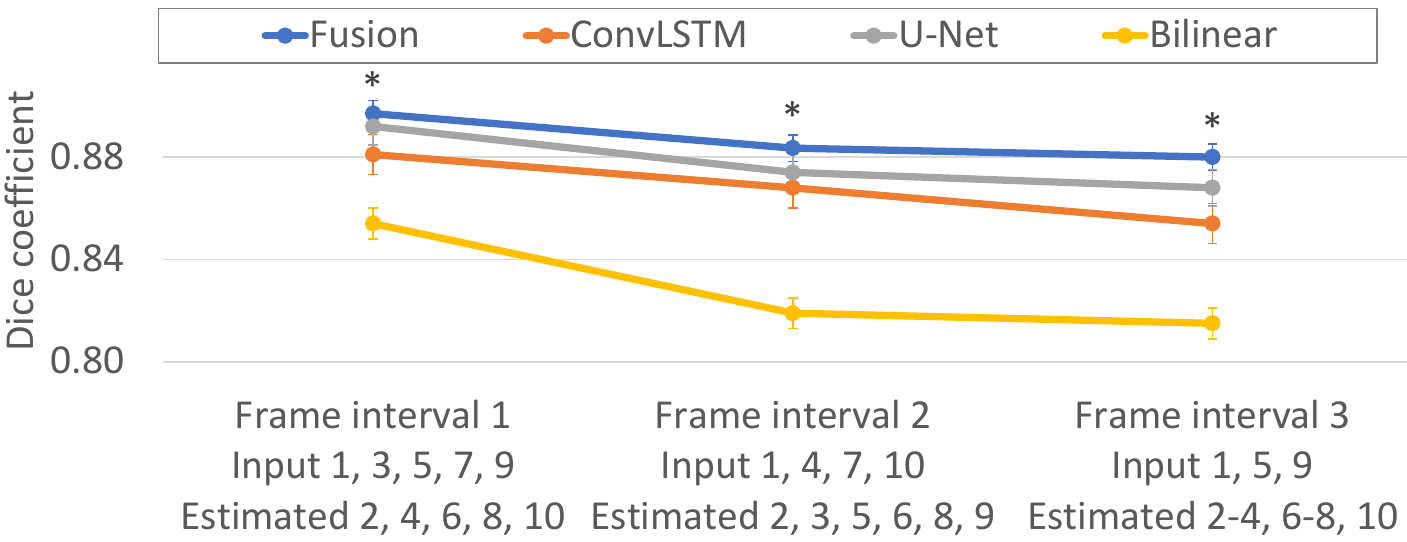}}
\caption{(a) Average volume change in ground truth 4D cardiac models. (b) Relationship between the average Dice coefficients and the input frame interval for FusionNet and the three existing methods (* p < .05).}
\label{fig:graph_of_frame_interval}
\end{center}
\end{figure}

Table \ref{table:comparison} shows that frames 2, 4, and 6 exhibit the lowest accuracies. Figure \ref{fig:graph_of_frame_interval} (a) shows the average volume change of the voxel models in each frame, which was calculated from all ground truth 3D voxel models by counting the number of voxels corresponding to the heart region within the model. It can be observed in this figure that frames 2, 4, and 6, which have low accuracy, correspond to frames with drastic volume changes around the end-systole. Even in such cases, the proposed FusionNet outperformed the existing methods in all frames.

Here, a left ventricular (LV) myocardium is treated as the heart region in this study.
Dice coefficients for manual LV myocardium segmentation results between different observers were reported to be 0.87-0.88 \cite{bai2018automated}.
In all frames, the Dice coefficients of the proposed FusionNet are equal to or exceed this inter-individual difference.
These results show that FusionNet provides stable shape estimation regardless of the frame.

\subsection{Ablation study}
We then conducted 7-fold cross-validation by comparing our complete FusionNet architecture with the following configurations: FusionNet
without skip connections (FusionNet-SC), FusionNet without residual blocks (FusionNet-RB), FusionNet without spatiotemporal encoders (FusionNet-TE), and the baseline described in Section \ref{sec:fusionnet}. 
The average Dice coefficients for each configuration were $\bf{0.897 \pm 0.019}$ (FusionNet), $0.810 \pm 0.015$ (FusionNet-SC), $0.891 \pm 0.019$ (FusionNet-RB), $0.892 \pm 0.018$ (FusionNet-TE), and $0.806 \pm 0.016$ (baseline).

Based on the average Dice coefficients, FusionNet exhibited the highest accuracy among the five configurations. 
This improvement was statistically significant at a significance level of 0.005.
The results showed that the addition of each element had a positive effect on FusionNet.

\subsection{Robustness to changes in frame intervals}
To evaluate the robustness of FusionNet to changes in the frame interval, 7-fold cross-validations were performed for the LFR heart models generated using three different frame intervals for the input: input frame interval 1 (input frames 1, 3, 5, 7, 9, and estimated frames 2, 4, 6, 8, and 10), input frame interval 2 (input frames 1, 4, 7, 10, and estimated frames 2, 3, 5, 6, 8, and 9), and input frame interval 3 (input frames 1, 5 ,9, and estimated frames 2, 3, 4, 6, 7, 8, and 10).

Figure \ref{fig:graph_of_frame_interval} (b) shows the generation accuracy for different networks using different frame intervals, indicating that FusionNet is statistically better at a significance level of 0.05.
Furthermore, FusionNet exhibits a slower decrease in accuracy with increasing frame intervals than the methods based on ConvLSTM and U-Net.
These results show that FusionNet, which introduces spatiotemporal convolution into a generative model, is more robust to changes in the frame interval than conventional RNN models or U-Net-based methods which only consider spatial convolution.

\section{Conclusion}
In this study, we propose FusionNet to estimate a HFR heart model from the corresponding LFR heart model.
The experimental results confirmed that FusionNet with spatiotemporal convolution was more accurate and robust to changes in the frame interval than conventional methods using RNNs or spatial convolution alone.

In future work, we can increase the sampling frequency but always need to keep the full cardiac cycle.
Also, to further improve the accuracy of frame interpolation, we plan to use not only a 4D heart model but also CMR images as inputs for FusionNet.
In addition, by applying FusionNet, we aim to develop a diagnostic support system for heart diseases that can provide highly accurate results, even from low-frame-rate CMR images.

\subsubsection{Acknowledgements}
This work was supported by JSPS KAKENHI Grant Number 20K19924, the Wallenberg AI, Autonomous Systems and Software Program (WASP), Sweden funded by the Knut and Alice Wallenberg Foundation, Sweden, 
and used the UK Biobank Resource under application no. 20209.

\bibliographystyle{splncs03}
\bibliography{miccai}

\end{document}